\documentclass[paper]{IEEEtran}
        
  	\usepackage[pdftex]{graphicx}
  	\usepackage{makecell}
  	\graphicspath{{../pdf/}{../jpeg/}}
	\DeclareGraphicsExtensions{.pdf,.jpeg,.png}
	\usepackage[cmex10]{amsmath}
	\usepackage{graphicx}
	\usepackage{algorithmic}
	\usepackage{array}
	\usepackage{mdwmath}
	\usepackage{mdwtab}
	\usepackage{eqparbox}
	\usepackage{url}
    \usepackage{float}
    \usepackage{afterpage}

\usepackage{xcolor}
\usepackage{accents}
\usepackage{amssymb}

\usepackage{adjustbox}

\begin{document}

\title{GHNet: Learning GNSS Heading from Velocity Measurements}
\author{Nitzan Dahan and Itzik Klein 
\thanks{Nitzan Dahan and Itzik Klein  are with the Hatter Department of Marine Technologies, Charney School of Marine Science, University of Haifa, Haifa, 3498838, Israel (e-mail: nitzandahan5@gmail.com, kitzik@univ.haifa.ac.il).}}
\maketitle

\begin{abstract}
By utilizing global navigation satellite system (GNSS) position and velocity measurements, the fusion between the GNSS and the inertial navigation system provides accurate and robust navigation information. When considering land vehicles, like autonomous ground vehicles, off-road vehicles, or mobile robots, a GNSS-based heading angle measurement can be obtained and used in parallel to the position measurement to bound the heading angle drift. Yet, at low vehicle speeds (less than 2m/s) such a model-based heading measurement fails to provide satisfactory performance. This paper proposes GHNet, a deep-learning framework capable of accurately regressing the heading angle for vehicles operating at low speeds. We demonstrate that GHNet outperforms the current model-based approach for simulation and experimental datasets.  
\end{abstract}

\IEEEoverridecommandlockouts

\IEEEpeerreviewmaketitle
\section{Introduction}\label{sec:intro}
An inertial navigation system (INS) is a dead reckoning (DR) navigation system that integrates the outputs of the inertial sensors to calculate the current position, velocity, and orientation of a platform \cite{britting2010inertial,titterton2004strapdown}. The INS is a self-contained navigation system that does not require external signals or communication. It consists of an inertial measurement unit (IMU) and a navigation processor. Because of errors in the inertial measurements, the solution of the INS diverges over time. To reduce the navigation solution drift, the INS is commonly fused with external sensors, like a global navigation satellite system (GNSS),
 \cite{noureldin2012fundamentals,farrell2008aided} or information about the environment \cite{klein2010pseudo,engelsman2023information} . \\
The fusion between the global navigation satellite system (GNSS) and the inertial navigation system (INS) is well known to fully utilize the advantages of the two systems to provide accurate and robust navigation information \cite{farrell2008aided}. Therefore, INS/GNSS fusion is highly researched topic in the last several decades including topics of the fusion type \cite{grewal2007global,groves2015principles,7004783}, system observability \cite{rhee2004observability,tang2008ins}, estimating the lever arm between the sensors \cite{seo2006lever,borko2018gnss}, and coping with situations of partial GNSS availability \cite{9722837,klein2011modified}.
Most of the literature focuses on the INS aided by the GNSS measured position and velocity vectors that can be provided using a single GNSS antenna. For a GNSS with dual antennas, heading information can be obtained using prior knowledge of the distance between the antennas \cite{Pereira2016GPSHA,wang2019research,roth2012improving,petovello2017you,pereira2016performance}. \\
When considering land vehicles, like autonomous ground vehicles, off-road vehicles, or mobile robots, heading information can also be obtained from the GNSS velocity measurements while using a single GNSS antenna. The calculated heading angle can then be used as an additional (to the position and velocity updates)  measurement update in the filter. In that case, the model-based approach of calculating the heading angle is done through an analytical trigonometric calculation of the GNSS velocity components. Such an approach was initially proposed in \cite{shin2002accuracy} for a velocity matching alignment approach. Later, this approach was implemented for kinematic azimuth alignment for land vehicles \cite{salycheva2004kinematic,godha2006performance}.\\
According to the paper \cite{shin2002accuracy,godha2006performance}, heading and roll angles are calculated using derived velocity values.
Another possibility to obtaining heading information using a single GNSS antenna was proposed in \cite{9069956} using a set of previous GNSS position measurements and a dynamic model while the vehicle is traveling in a straight line trajectory allowed the estimation of the heading angle.\\
The main shortcoming of the model-based GNSS heading is its poor performance at low vehicle speeds. The reason is a low signal-to-noise ratio as well as numerical issues with small values in the inverse trigonometric function. Motivated by the successful integration of the deep-learning approach applied to many different navigation tasks \cite{klein2022data,cohen2023inertial}, this research proposes GHNet, a learning framework to accurately regress the heading angle at low vehicle speeds using a single antenna. GHNet provides the heading angle utilizing the GNSS velocity measurement at each epoch without relying on past GNSS measurements. We first highlight the model-based inability to estimate the heading angle at a low speed and then derive and present our approach. We evaluate the proposed approach using simulations and experiment data. To that end, we utilize the public dataset recorded by the University of Michigan North Campus Long-Term Vision and LiDAR (NCLT), which contains Segway recordings with a duration of 321 minutes. The dataset includes different driving speeds while driving outside and inside buildings. Our results show that using the proposed approach, a GNSS based heading angle can be used even at low vehicle speeds. \\
The rest of the paper is organized as follows: Section II presents the traditional heading angle  approach. Section III gives the proposed GHNet approach. Section IV presents the datasets. Section V presents the analysis and results while Section VI gives the conclusions of the paper.
\section{GNSS-Based Heading Angle}\label{sec:gnssheading}
\begin{figure}[ht!] 
    \centering
    \includegraphics[width=1.5in]{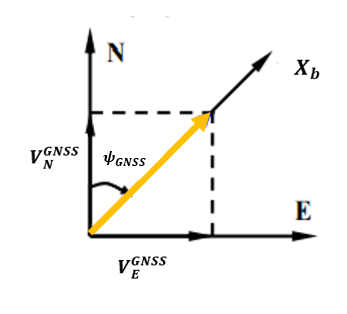}
    \caption{GNSS-based heading problem geometry.} \label{fig:heading1}
\end{figure}

\begin{figure}[ht!] 
    \centering
    \centerline{\includegraphics[width=\columnwidth]{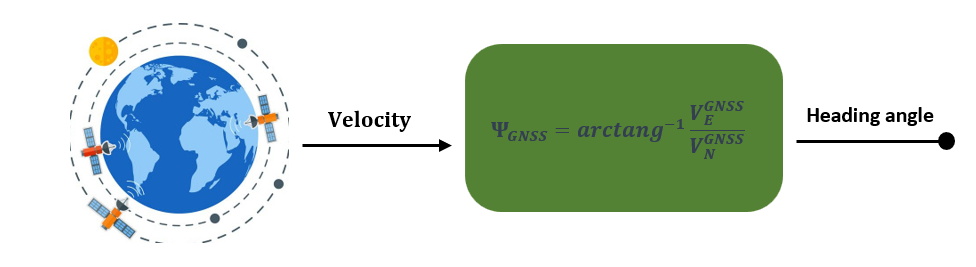}}
    \caption{Overview of the model-based heading estimation approach.} \label{fig:classic_approach}
\end{figure}

In addition to the position and velocity vectors, heading information can be extracted from the GNSS velocity measurements when considering land vehicles or mobile robots moving in the horizontal plane as presented in Figure~\ref{fig:heading1}.  
From the problem geometry, the heading angle is defined as \cite{base_navigation_equation}:
 \begin{equation}\label{eq:psi_gnss}
    \psi_{GNSS} = \arctan{\frac{v_E^{GNSS}}{v_N^{GNSS}}}
\end{equation}
where $\psi_{GNSS}$ is the GNSS based heading angle, and $v_E^{GNSS}$ and $v_N^{GNSS}$ are the GNSS velocity components in east and north directions, respectively. 
We refer to this heading estimation approach (reflected in \eqref{eq:psi_gnss}) as the model-based approach and provide an illustration of the approach in Figure~\ref{fig:classic_approach}.\\
The standard deviation of the heading error, $\sigma_\psi$, is obtained using  \eqref{eq:psi_gnss}
 \begin{equation}\label{eq:heading_error_equ}
   {\sigma_\psi}=\frac{\sigma (v) }{v^{GNSS}}
\end{equation}
where $\sigma_v$ is the standard deviation of the GNSS velocity measurement and $v^{GNSS}$ is the horizontal velocity vector magnitude defined by
 \begin{equation}\label{eq:v_gnss}
    v^{GNSS} =\sqrt{(v_E^{GNSS})^2+(v_N^{GNSS})^2}.
\end{equation}
Thus, the accuracy of the heading angle error is determined by the accuracy of the GNSS velocity measurements  divided by the vehicle speed.
As a consequence, as the speed of the vehicle decreases, the accuracy of the heading estimation decreases until a certain point where the heading estimation is pointless. 
\section{Proposed Approach}\label{sec:pr}
A motivation for our proposed approach is given, followed by a description of our learning method and network structure. 
\subsection{Motivation}
\begin{figure}[ht!] 
    \centering     
    \includegraphics[width=0.93\linewidth]{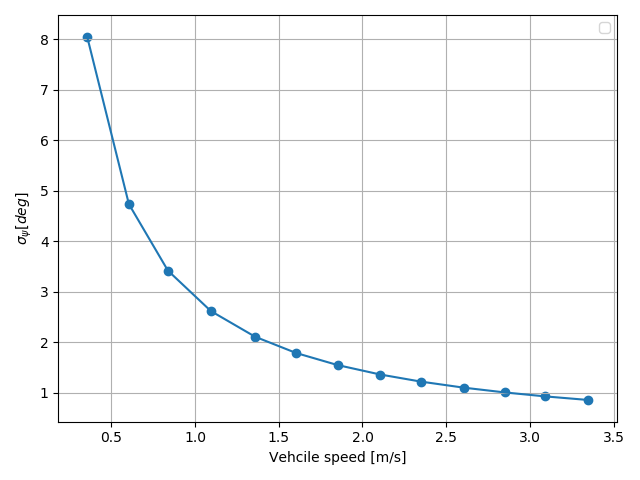}
    
    \caption{The standard deviation of the heading angle as a function of the vehicle speed for the simulated example.}
    \label{fig:sig_vs_speed}
\end{figure}

When a vehicle is moving at low speeds, the heading estimation based on GNSS signals is challenging. To demonstrate this claim, we consider the following scenario: a vehicle  traveling with different speeds ranging from $0. 1-3.5$ m/s.  We define the body coordinate frame such that its  center is located at the center of mass, the $x$-axis is parallel to the longitudinal axis of symmetry of the vehicle pointing forward, the $y$-axis points right, and the $z$-axis points down such that it forms a right-handed orthogonal frame. 
In land vehicles, given no-slip conditions, the velocity components perpendicular to the forward direction, y-and z-axes, are assumed to be zero and the vehicle speed is equal to the velocity vector component in the x-axis. That is,
\begin{equation}\label{eq:velb}
    \mathbf{v}^{b} =  \begin{bmatrix} v_x  & 0 & 0  \end{bmatrix}^T.
\end{equation}
As the GNSS velocity measurements are given in the navigation frame, a transformation between the body and navigation frames is needed.  Such a transformation matrix is defined by
\begin{equation}\label{eq:tbn}
   \mathbf{T}_b^n=\begin{bmatrix}
c\psi c\theta& -s\psi c\phi +c\psi s\theta s\phi & s\psi s\phi +c\psi c\phi s\theta\\
s\psi c\theta & c\psi c\phi +s\phi s\theta s\psi & -c\psi s\phi + s\theta s\psi c\phi\\
-s\theta & c\theta s\phi & c\theta c\phi
\end{bmatrix}
\end{equation}
where $\phi$ is the roll angle, $\theta$ is the pitch angle, and $\psi$ is the heading angle. \\
For demonstration purposes, without loss of generality, it is assumed that the roll and pitch angles are zero. 
In addition, we arbitrarily chose an heading angle of $\psi=60$ deg. Thus, the transformation matrix \eqref{eq:tbn} reduces to:
\begin{equation}\label{eq:tbn1}
   \mathbf{T}_b^n=\begin{bmatrix}
c\frac{\pi}{3} & -s\frac{\pi}{3}  & 0\\
s\frac{\pi}{3}  & c\frac{\pi}{3}  & 0\\
0 & 0 & 1
\end{bmatrix}.
\end{equation}
Using \eqref{eq:tbn1}, the velocity vector in the navigation frame is obtained by
 \begin{equation}\label{eq:vn}
   {\mathbf{v}^{n}}=\mathbf{T}_b^n\cdot{\mathbf{v}^{b}}
\end{equation}
where $\mathbf{v}_{n}$ is the velocity in the navigation frame and $\mathbf{v}_{b}$ is the velocity vector in the body frame. \\
To simulate GNSS measurements, we define the value of the vehicle speed \eqref{eq:velb}, calculate the velocity vector in the navigation frame using \eqref{eq:vn}, and add a zero white Gaussian noise to the velocity vector. The measured velocity vector is plugged into (2) to calculate the heading STD. Next, the heading STD as a function of the speed was generated and plotted in Figure~\ref{fig:sig_vs_speed}. It can be observed that the standard deviation of the heading angle decreases as the speed value increases. Yet, at low speeds, the error distribution of the heading angle dramatically increases,  leading to large errors in heading angle estimation.\\
To overcome this shortcoming, we are looking for a method that enables accurate heading estimation even at low vehicle speeds.
\subsection{GHNet Framework}\label{sec:pr}
\begin{figure}[ht!] 
    \centering
    \centerline{\includegraphics[width=\columnwidth]{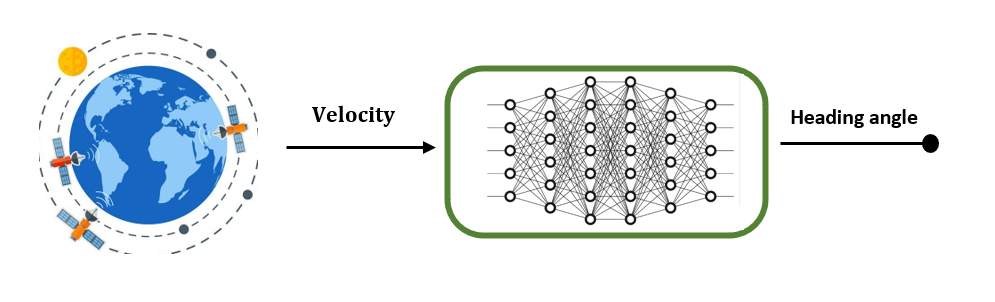}}
    \caption{GHNet learning the heading angle from GNSS velocity measurements.} \label{fig:gnss_approach}
\end{figure}

\begin{figure*}[ht!] 
\centering
\includegraphics[width=6in]{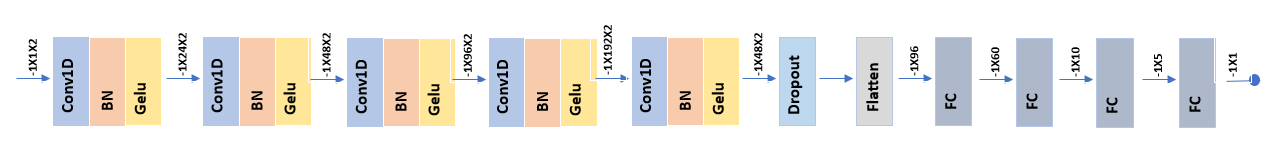}
\caption{GHNet baseline neural network structure has five convolutional layers, followed by dropout and flatten layers, and  four linear layers. }
\label{fig:CNN_block}
\end{figure*}

The goal of this paper is to derive an end-to-end GNSS heading network, to regress the heading angle given the GNSS velocity measurements. Specifically, the GHNet goal is to enable accurate heading estimation even at low vehicle speeds. Figure~\ref{fig:gnss_approach} illustrates our proposed GHNet approach. The main idea of using a network instead of the analytical model-based is to benefit from well-known deep-learning capabilities such as noise reduction and to capture nonlinear behavior, particularly in situations of a low single-to-noise ratio (as in our problem).  In this manner, the corresponding heading STD decreases, thus allowing accurate heading estimation even when the vehicle moves at low speeds.\\
From a machine learning point of view, we address a supervised regression problem. That is, the regression model is a process to model the relationship between the input (GNSS velocity) and output (heading angle) variables. To solve such a regression problem, we employ neural network approaches. 
Our deep network baseline architecture for GHNet is presented in Figure~\ref{fig:CNN_block}. The network receives GNSS measured velocity in the north and east components as input (same as the model-based approach), and in its last layer, predicts the heading angle. The size and activation function of each layer are given in Table ~\ref{table:proposed_arch}.
\begin {table}[H]
\caption {Proposed network architecture layers, type, and activation functions.}
\begin{center}
\begin{tabular}{|c c c c|} 
 \hline
 Layer & Type & Size & Activation \\ [0.5ex] 
 \hline
 0 & Input & -1*1*2 & -\\ 
 1 & Conv1D &1*1,24 & Gelu\\
 2 & Conv1D &1*1,48  & Gelu\\
 3 & Conv1D &1*1,96  & Gelu\\
 4 & Conv1D &1*1,192  & Gelu\\
 5 & Conv1D &1*1,48  & Gelu\\
 6 & Dropout & - & -\\
 7 & Flatten & - & -\\
 8 & FC & 60  & -\\
 9 & FC & 10   & -\\
 10 & FC & 5   & -\\
 11 & FC & 1   & -\\
 [1ex] 
 \hline
\end{tabular}
\label{table:proposed_arch}
\end{center}
\end {table}
Our model is structured in the following layers:
\begin{itemize}
  \item \textbf{1D-CNN layer}: A one-dimensional convolutional neural network layer is the main building block of the baseline network. It consists of filters to extract features from the input signal and kernels to specify the height of the filter. During the training phase, the 1D-CNN learns the optimal weights of the kernels. The input layer is followed by a chain of five 1D-CNN layers. There are 24 kernels for the first layer, 48 kernels for the second layer, 96 kernels for the third layer, 192 kernels for the fourth layer, and 48 kernels for the fifth layer. \\
  A general convolution 1D layer equation is defined as
\begin{equation}
     \textbf{y}_{i\textquotesingle c\textquotesingle} = \sum_{ic} \textbf{W}_{icc\textquotesingle}\textbf{x}_{i+i\textquotesingle,c}+\textbf{b}_{c\textquotesingle}
     \label{eq:conv_equ}
\end{equation}
    where $\textbf{x}$ represents the input to the convolutional function, $\textbf{W}$ represents the filters, and $\textbf{y}$ represents the output of the convolutional layer. The data $\textbf{x}$ has dimensions of M × C, where M represents the number of latent features and C is the number of channels.
  \item \textbf{Batch normalization layer}: A major challenge of 1D-CNN is that its gradients are highly dependent on the output of neurons in the previous layer, especially if these outputs change in a highly correlated manner \cite{Batch1}. Batch normalization, and in particular layer normalization, is proposed to reduce such an undesirable covariate shift, stabilize the learning process, and significantly reduce the number of training epochs for model \cite{Batch2}. We added a batch normalization layer after every 1D-CNN layer using this definition: 
    \begin{equation}
         \textbf{y}^n = ({\frac{\textbf{y}-\textbf{m}_y}{\textbf{s}_y}})\gamma+ \boldsymbol{\beta}
        \label{eq:Batchnormalization_equ}
    \end{equation}
    where $\textbf{y}^n$ represents the output of the batch normalization layer, $\textbf{m}_y$ is the mean of the neurons output, $\textbf{s}_y$ is the standard deviation of the output of the neurons, and $\gamma$ and $\beta$ are the learning parameters of the batch normalization layer.
 \item \textbf{GELU function}: An activation function of the Gaussian error linear unit is proposed by Hendrycks and Gimpel \cite{Gelus2}:
    \begin{equation}
         GELU(x) = 0.5x(1+tanh[\sqrt{\frac{2}{\pi}}(x+0.044715x^3)])
        \label{eq:gelu_equ}
    \end{equation}
    where x is the input feature.
    The GELU function represents nonlinearity using the stochastic order in the input. This function takes advantage of the negative value. When the values are positive, the behavior of GELU is the same as that of ReLU; however, when the values are negative, the gradient value is positive and in this range GELU has a clear advantage over ReLU. 
    In the case the value are negative, as with ReLU, the gradient value is zero\cite{Gelus3}. For these reasons, the GELU activation functions are employed after each 1D-CNN layer in our baseline network architecture.
 \item \textbf{Dropout layer}: The dropout layer randomly assigns zero weights to neurons of the model to make it less sensitive to smaller variations, thereby improving the accuracy of the model on unseen data and preventing the models from over fitting \cite{Droput2}.\\Our model uses this layer after the last convolutional layer. As shown in Table ~\ref{table:proposed_arch}, this is the sixth layer.
  \item \textbf{Flatten layer}: Converts the 2D featured map matrix to a 1D feature vector and allows the output to be handled by the following fully connected layers.
    Our model uses this layer after the dropout layer, where the features are flattened by stacking the individual vectors on top of each other. 
    As shown in Table ~\ref{table:proposed_arch}, this is the seventh layer.
    \item \textbf{Fully connected layer}: Linear transformation is applied to the incoming data. Input features are received as flattened 1D vectors and multiplied by a weighting matrix.  Our model has four fully connected layers. There are sixty kernels for the first layer, ten kernels for the second layer, five kernels for the third layer, and one kernel for the fourth layer. The fully connected layer equation is given by 
    \begin{equation}
         \textbf{y}_{i\textquotesingle} = \sum_{i} \textbf{W}_{ii\textquotesingle}\textbf{x}_{i}+\textbf{b}_{i\textquotesingle}
     \label{eq:fc_equ}
    \end{equation}
    where x represents the input to the fully connected layer, W represents the weight matrix, b is the bias, and y represents the output of the fully connected layer.
 \item \textbf{Loss function}: The mean absolute error (MAE) is employed as the loss function . The MAE is defined by
    \begin{equation}
        \mathcal{L}_{MAE}= \frac{1}{N} \sum\limits_{j=1}^N |{\textbf{h}_j -{\mathbf{\tilde{h}}(x,y)}_j}|
        \label{eq:mae}
    \end{equation}
    where $N$ is the number of samples, $\mathbf{\tilde{h}}$ is the predicted value, and $\textbf{h}$ is the true value. In our model, $\mathbf{\tilde{h}}$ is the predicted heading angle and $\textbf{h}$ is the true heading angle.
\end{itemize}
AdamW \cite{adamW} is selected as the optimizer for the proposed model and the dropout layer is set with a value of 0.1. The GHNet architecture is shown in Figure ~\ref{fig:CNN_block} and consists of 70,709 parameters.
\section{Datasets}\label{sec:pr}
Two different datasets were employed in this research for the training dataset. The first is a simulation dataset used for initial evaluations followed by the NCLT Segway dataset~\cite{NCLT}. The test dataset consists of only NCLT recordings. 
\subsection{Simulative Dataset}
\begin{figure}[ht!] 
    \centering
    \centerline{\includegraphics[width=\columnwidth]{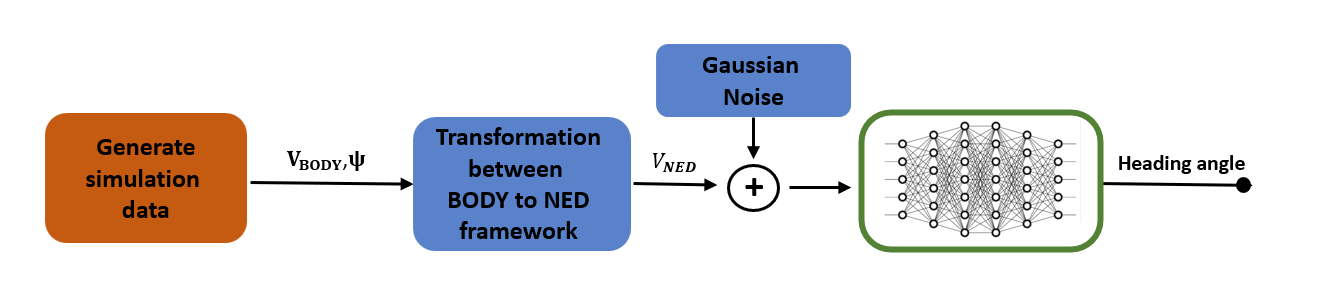}}
    \caption{Overview of data generation for simulation data for evaluating GHNet.} \label{fig:approch_sim}
\end{figure}

To evaluate our proposed approach, we created a simplified simulation based dataset. The flowchart of generating the data is illustrated in Figure ~\ref{fig:approch_sim}. In the simulation, the user determines the true vehicle speed and heading angle, while the roll and pitch angles are assumed to be zero. As the vehicle velocity vector in the body frame is created, it is transformed using \eqref{eq:tbn} to the navigation frame to obtain the nominal GNSS velocity vector. To generate GNSS data, zero mean white Gaussian noise with STD of $0.01$m/s is added to the velocity components. 
The above procedure was repeated several times with different parameters. The vehicle speed was in the range of $0-3.5$ m/s and the heading angle range was $0-90$ deg with intervals of $0.1$ and $0.5$, respectively. Each combination of speed and angle was repeated ten times, each with different noise values. In total, $7,317$ samples were generated and divided into train and validation sets using a random $80 -20\%$ split. Table ~\ref{table:sim_dataset} summarizes the simulation dataset.
\begin {table}[H]
\caption {Summary of simulative dataset properties.}
\begin{adjustbox}{width=.5\textwidth,center}
\begin{tabular}{|c c c c c|} 
 \hline
Data label &Name &Num of  &Speed&Heading \\ 
 & &samples&range &range \\
&&&[m/s] &[deg] \\[0.5ex]
 \hline
  S1 & Simulative dataset & 7,317 & 0:3.5 & 0:90 \\ [1ex] 
 \hline
\end{tabular}
\label{table:sim_dataset}
\end{adjustbox}
\end {table}
\subsection{NCLT Dataset}\label{sec:ncltd}
The University of Michigan’s North Campus Long-Term Vision and LiDAR (NCLT) dataset is a commonly used, publicly available  dataset for robotics research~\cite{NCLT}. The NCLT dataset contains different sensors mounted on Segway robots moving in different types of dynamics and trajectories, both indoors and outdoors. Form this dataset, we took the ground truth trajectory of the velocity and heading angles as well as the noisy GPS measurements. \\
For the construction of our train and validation datasets, we randomly selected two dates of recordings; namely, 2012-01-15 and 2012-02-05. In these recordings, the heading angle range is $0-90$ [deg] and the vehicle speed is greater than or equal to zero (stationary conditions are not considered).
Following the same procedure as in the simulation dataset, this data was randomly divided into training and validation sets with an $80 -20\%$ ratio. 
Table ~\ref{table:NCLT_train_dataset} shows our selection of the NCLT train datasets with their properties. In total, there are $147,367$ samples recording at  a frequency of $10$Hz, corresponding to approximately 245 minutes of data.
\begin {table}[H]
\caption {Summary of NCLT train dataset properties.}
\begin{adjustbox}{width=.5\textwidth,center}
\begin{tabular}{|c c c c c c|} 
 \hline
 Data label &Name &Num of  &Speed&Heading&Time \\ 
 & &samples&range &range& \\
&&&[m/s] &[deg]&[m] \\[0.5ex]
 \hline
 E1 & 2012-01-15 & 85,429 & 0.01:2.16& 0:90 &142\\ 
 E2 & 2012-02-05 & 61,938 & 0.01:2.2 &0:90 &103\\
 E3 & E1+E2 & 147367 & 0.01:2.2 & 0:90 &245\\[1ex]
 \hline
\end{tabular}
\label{table:NCLT_train_dataset}
\end{adjustbox}
\end {table}

For the test dataset, we chose another three random dates: 2012-01-08, 2012-02-02, and 2012-11-17. Table ~\ref{table:NCLT_test_dataset} details the test dataset.  In total,  321 minutes of recordings were used in the test datasets.
\begin {table}[H]
\caption {Summary of NCLT test dataset properties.}
\begin{adjustbox}{width=.5\textwidth,center}
\begin{tabular}{|c c c c c c|} 
 \hline
Data label &Name &Num of  &Speed&Heading&Time \\ 
 & &samples&range &range& \\
&&&[m/s] &[deg]&[m] \\[0.5ex]
 \hline
T1& 2012-01-08 & 64,695 & 0:2 & 0:90 &108\\
T2& 2012-02-02 & 72,109 & 0:2 & 0:90 &120\\
T3& 2012-11-17 & 55,903 & 0:2 & 0:90 &93\\
T4& T1+T2+T3 & 192,717 & 0:2 & 0:90 &321\\[1ex]
 \hline
\end{tabular}
\label{table:NCLT_test_dataset}
\end{adjustbox}
\end {table}
\section{Analysis and Results}\label{sec:pr}
We begin this section by defining our evaluation metrics followed by simulation and experiment results. 
\subsection{Evaluation Metrics}
Two evaluation metrics are considered:
\begin{itemize}
    \item \textbf{Mean absolute error (MAE)}: Computes the mean absolute error between the ground truth value and its estimated value: 
        \begin{equation}
        MAE= \frac{1}{N} \sum\limits_{j=1}^N |{{h}_j -{{\tilde{h}}}_j} |
        \label{eq:mae_evaluation}
    \end{equation}
    
    \item \textbf{Root mean square error (RMSE)}: Computes the root of the average of the squared difference between the ground truth and its estimated counterpart, defined by
    \begin{equation}
         RMSE = \sqrt{{\frac{1}{N}}\sum\limits_{j=1}^N({h_j - {\tilde{h}}_j)^2}}
        \label{eq:rmse}
    \end{equation}
    where $N$ is the number of samples, $\tilde{h}$ is the estimated value, and $h$ is its ground truth value. In our analysis, $\tilde{h}$ is the predicted heading angle and $h$ is its true heading angles.
\end{itemize}
\subsection{Simulation results}
\begin{figure}[ht!] 
    \centering
    \centerline{\includegraphics[width=\columnwidth]{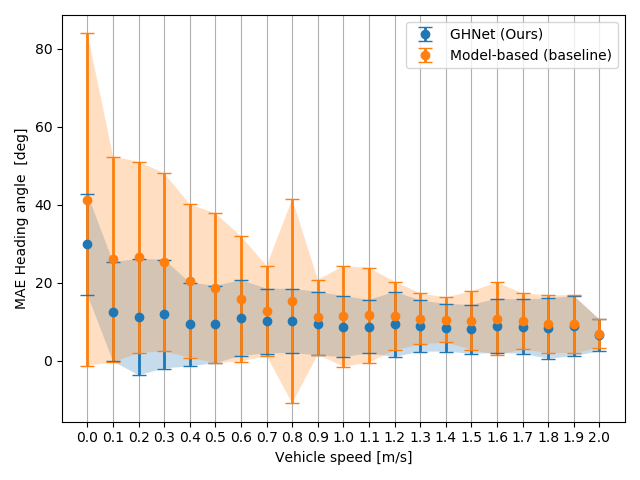}}
    \caption{MAE results of GHNet and model-based. The shadow of each curve presents the STD.} \label{fig:preformances_sim}
\end{figure}

\begin{figure}[ht!] 
    \centering
    \centerline{\includegraphics[width=\columnwidth]{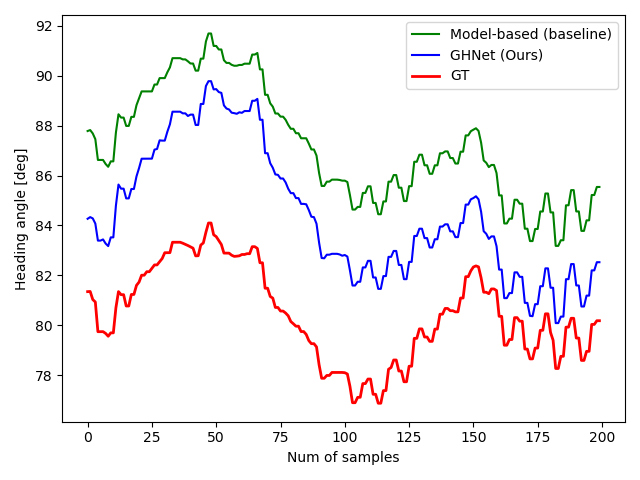}}
    \caption{Heading angle prediction results for an arbitrary set of 200 consecutive samples from the test dataset.} \label{fig:1100myplot1}
\end{figure}

\begin{figure}[ht!] 
    \centering
    \centerline{\includegraphics[width=\columnwidth]{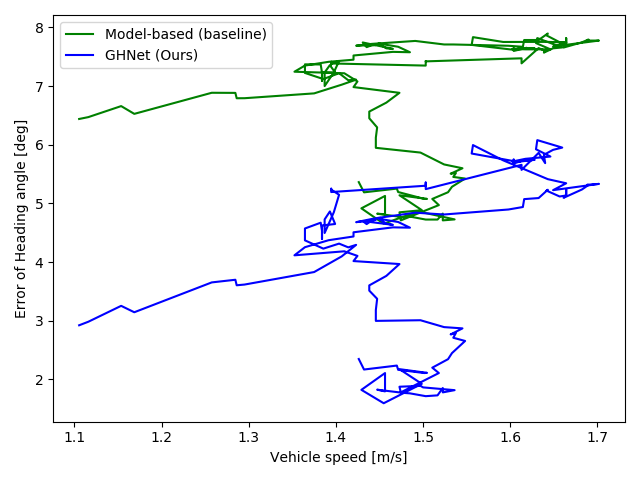}}
    \caption{Heading angle prediction results as a function of the vehicle speed for an arbitrary set of 200 consecutive samples from the test dataset.} \label{fig:myplot5}
\end{figure}

The simulation dataset was used for two purposes: 1) validating our  GHNet approach and 2) as the train dataset for the NCLT test dataset. The validation on the train simulation data obtained excellent performance and is not given here. We focus on the case where the simulation generated dataset, S1, is used as the train dataset while the NCLT T4 set is used for testing. That is, we explore the possibility of using only simulation to train the network instead of extensive and costly field experiments.\\
The results, divided by intervals of $0.1$m/s, are presented in Table~\ref{table:result_sim_model}. The table gives both the RMSE and MAE metrics for the model-based (baseline) and our proposed GHNet network, as well as the improvement of GHNet compared to the model-based approach. The results in the table demonstrate that our GHNet network, when trained on a simulation dataset, improves the model-based approach up to vehicle speeds of $2$m/s. For higher vehicle speeds, the baseline approach is preferred. Specifically, the speed range of $0.01 -0.8$ m/s obtains an average improvement of 41.8\%. When the vehicle speed is increased to the range of $0.9-2$m/s, the improvement rate is 16.3\%.
As a point of clarification, all the improvements noted above are based on the MSE.
\begin{table}[]
\caption {Model-based (MB) and our GHNet performance on a simulation trained network (dataset S1) and tested on the NCLT test dataset T4.}
\begin{adjustbox}{width=.5\textwidth,center}
\begin{tabular}{|ccccccc|}
\hline
Speed & &RMSE [deg] &&&MAE [deg]& \\
range [m/s] &MB &GHNet  &Impro-&MB&GHNet&Impro- \\ 
&(baseline) &(ours)&vement\%\ &(baseline)&(ours)&vement \%\ \\[0.5ex]
\hline
\textless{}0.1 & 59.3 & 32.5 & 45.2 & 41.3 & 29.8 & 27.8 \\
0.1 & 36.9 & 17.9 & 51.5 & 25.9 & 12.6 & 51.4 \\
0.2 & 36.1 & 18.6 & 48.5 & 26.5 & 11.2 & 57.7 \\
0.3 & 34.1 & 18.4 & 46 & 25.2 & 11.9 & 52.7 \\
0.4 & 28.3 & 14.2 & 49.8 & 20.4 & 9.3 & 54.4 \\
0.5 & 26.7 & 13.6 & 49.1 & 18.6 & 9.3 & 50 \\
0.6 & 22.5 & 14.6 & 35.1 & 15.8 & 11.1 & 29.7 \\
0.7 & 17.2 & 13.1 & 23.8 & 12.7 & 10.2 & 19.7 \\
0.8 & 30.3 & 13.1 & 56.7 & 15.3 & 10.2 & 33.3 \\
0.9 & 14.7 & 12.5 & 14.9 & 11.2 & 9.5 & 15.2 \\
1 & 17.2 & 11.7 & 31.9 & 11.4 & 8.7 & 23.7 \\
1.1 & 16.8 & 11.1 & 33.9 & 11.6 & 8.7 & 25 \\
1.2 & 14.4 & 12.5 & 13.2 & 11.5 & 9.3 & 19.1 \\
1.3 & 12.6 & 11.1 & 11.9 & 10.7 & 8.8 & 17.7 \\
1.4 & 11.9 & 10.4 & 12.6 & 10.5 & 8.5 & 19 \\
1.5 & 12.7 & 10.2 & 19.7 & 10.3 & 8.1 & 21.4 \\
1.6 & 14.2 & 11.3 & 20.4 & 10.7 & 8.9 & 16.8 \\
1.7 & 12.4 & 11.2 & 9.7 & 10.1 & 8.7 & 13.8 \\
1.8 & 11.9 & 11.3 & 5 & 9.4 & 8.3 & 11.7 \\
1.9 & 11.9 & 11.7 & 1.7 & 9.4 & 8.8 & 6.4 \\
2 & 7.8 & 7.6 & 2.6 & 6.9 & 6.5 & 5.8 \\ \hline
\end{tabular}
\label{table:result_sim_model}
\end{adjustbox}
\end{table}
Figure ~\ref{fig:preformances_sim} presents, besides the MAE values, the corresponding STDs of the model-based and our GHNet approaches as a function of the vehicle speed. The figure shows that our proposed approach also manages to improve the STD compared to the model-based approach. \\
To further analyze the results, we took an arbitrary set of 200 consecutive samples from the test dataset.
Figure ~\ref{fig:1100myplot1} presents the heading angle prediction results of the baseline and  GHNet approaches as well as the ground truth (GT) values. As can be seen, our approach improves the baseline estimation in all epochs. 
To include the vehicle speed, in Figure~\ref{fig:myplot5} we show the model-based and GHNET heading angle prediction errors as a function of the vehicle speed. The plots show a significant improvement of GHNET throughout the entire speed range, and the error of GHNET is on average 1.8 times smaller than the error of baseline.
\subsection{Experiment results}
This section assesses the performance of our proposed approach while training on the NCLT dataset—described in Section~\ref{sec:ncltd}—and testing on the NCLT test dataset T4. Following in the same manner as for our simulation results, This table ~\ref{table:result_real_model} introduces the results in a similar manner than this table ~\ref{table:result_sim_model}. The table gives both the RMSE and MAE metrics for the model-based (baseline) and our proposed GHNet network, as well as the improvement of GHNet over the model-based approach. The table shows that the GHNET approach significantly improves the performance compared to the baseline for the entire range of speeds. Looking at the MAE metric, for the $0.01-0.8$ m/s speed range, an average improvement of 58\% was achieved compared to the baseline, while for the range of $0.9-2$ m/s an improvement of 61.8\% was achieved. As expected, when using the NCLT dataset as the training dataset, the performance improved compared to using a simulation dataset. Particularly the speed range of $0.9-2$ m/s achieved an average  improvement by a factor of 2.9. 
\begin{table}[]
\caption {Model-based (MB) and our GHNet performance on the NCLT trained network (dataset E3) and tested on the NCLT test dataset T4.}
\begin{adjustbox}{width=.5\textwidth,center}
\begin{tabular}{|ccccccc|}
\hline
Speed & &RMSE [deg] &&&MAE [deg]& \\
range [m/s] &MB &GHNet  &Impro-&MB&GHNet&Impro- \\ 
&(baseline) &(ours)&vement\%\ &(baseline)&(ours)&vement \%\ \\[0.5ex]
 \hline
\textless{}0.1 & 59.3 & 28.6 & 51.7 & 41.3 & 23.2 & 43.8 \\
0.1 & 37.1 & 16.2 & 56.3 & 25.9 & 10.1 & 61 \\
0.2 & 36.1 & 16.5 & 54.3 & 26.5 & 10.1 & 61.8 \\
0.3 & 34.1 & 16.3 & 52.2 & 25.2 & 9.7 & 61.5 \\
0.4 & 28.3 & 12.3 & 56.5 & 20.4 & 7.4 & 63.7 \\
0.5 & 26.7 & 10.8 & 59.5 & 18.6 & 6.8 & 63.4 \\
0.6 & 22.5 & 11.4 & 49.3 & 15.8 & 7.1 & 55.1 \\
0.7 & 17.2 & 10.4 & 39.5 & 12.7 & 6.1 & 51.9 \\
0.8 & 30.3 & 10.1 & 66.6 & 15.3 & 6.1 & 60.1 \\
0.9 & 14.7 & 7.6 & 48.3 & 11.2 & 4.7 & 58 \\
1 & 17.2 & 6.8 & 60.4 & 11.4 & 4.3 & 62.3 \\
1.1 & 16.8 & 5.6 & 66.6 & 11.6 & 3.7 & 68.1 \\
1.2 & 14.5 & 7.3 & 49.6 & 11.5 & 4.3 & 62.6 \\
1.3 & 12.5 & 6.6 & 47.2 & 10.7 & 4.1 & 61.7 \\
1.4 & 11.9 & 4.5 & 62.2 & 10.5 & 3.1 & 70.5 \\
1.5 & 12.7 & 4.1 & 67.7 & 10.3 & 2.7 & 73.8 \\
1.6 & 14.2 & 5.5 & 61.3 & 10.7 & 3.3 & 69.2 \\
1.7 & 12.4 & 6.6 & 46.8 & 10.1 & 4.1 & 59.4 \\
1.8 & 12 & 7.6 & 36.6 & 9.4 & 4.2 & 55.3 \\
1.9 & 12 & 7.4 & 38.3 & 9.4 & 4.3 & 54.3 \\
2 & 7.8 & 4.8 & 38.5 & 6.9 & 3.7 & 46.4 \\ \hline
\end{tabular}
\label{table:result_real_model}
\end{adjustbox}

\end{table}
\par
\begin{figure}[ht!] 
    \centering
    \centerline{\includegraphics[width=\columnwidth]{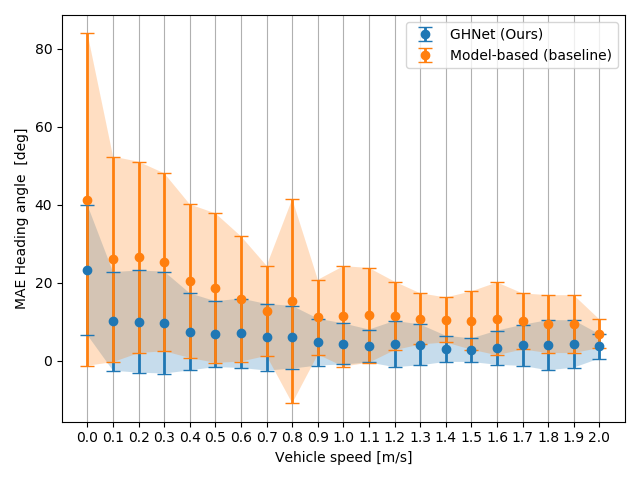}}
    \caption{MAE results of GHNet and model-based. The shadow of each curve presents the STD.} \label{fig:preformances_real}
\end{figure}

Figure~\ref{fig:preformances_real} presents, besides the MAE values, the corresponding STDs of the model-based and our GHNet approaches as a function of the vehicle speed. The figure shows that our proposed approach also manages to improve the STD compared to the model-based approach.
\par
\begin{figure}[ht!] 
    \centering
    \centerline{\includegraphics[width=\columnwidth]{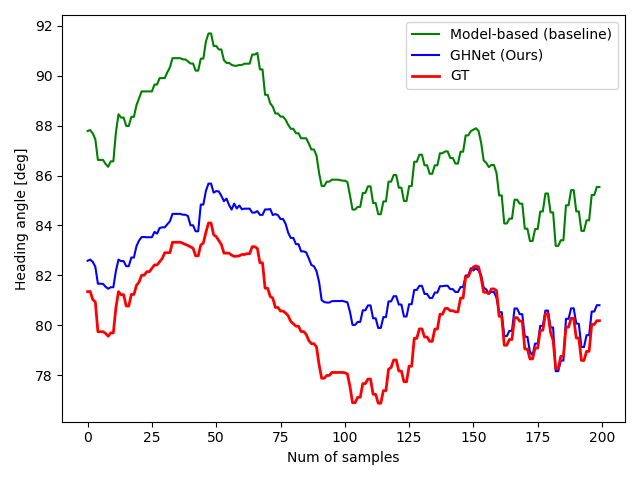}}
    \caption{Heading angle prediction results for an arbitrary set of 200 consecutive samples from the NCLT test dataset.} \label{fig:1100myplot1_real}
\end{figure}
\begin{figure}[ht!] 
    \centering
    \centerline{\includegraphics[width=\columnwidth]{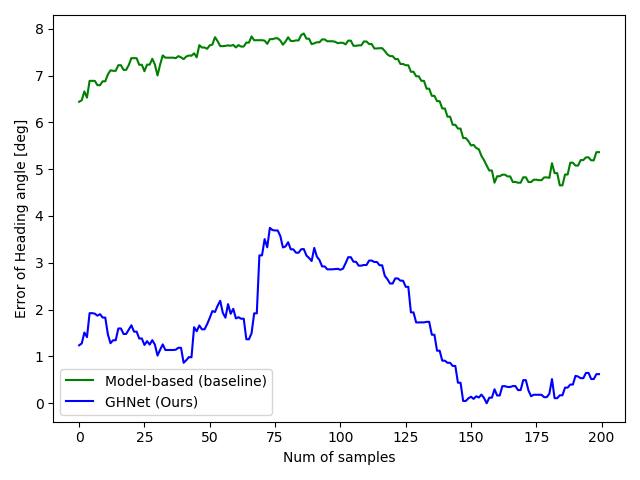}}
    \caption{Heading angle prediction results as a function of the vehicle speed for an arbitrary set of 200 consecutive samples from the test dataset.} \label{fig:1100myplot3_real}
\end{figure}

Next, we took an arbitrary set of 200 consecutive samples from the test dataset.
Figure ~\ref{fig:1100myplot1_real} presents the heading angle prediction results of the baseline and  GHNet approaches as well as the ground truth (GT) values. As can be seen, our approach improves the baseline estimation in all epochs. 
To also include the vehicle speed, Figure~\ref{fig:1100myplot3_real} shows the baseline and GHNET heading angle prediction errors as a function of the vehicle speed. The plots show a significant improvement of GHNET throughout the entire speed range, and the error of GHNET improves, on average, by 76.2\%. \\
Table~\ref{table:sumarry_results_general} summarizes the results of the baseline approach compared to the GHNet approach for four different test datasets, T1, T2, T3, and T4, as defined in Section \ref{sec:ncltd}. 
Here, GHNet is trained on both the simulation dataset S1 and the NCLT train dataset E3. Regardless of the training dataset, our GHNet outperformed the baseline approach. When trained on the simulative dataset, an average improvement of $27\%$ was achieved, while training on the experimental dataset an average improvement of $65\%$ was obtained. This shows that training on simulation data alone can improve the baseline algorithms. 
\begin{table}[]
\caption {Summary of the comparison between the baseline and GHNet approaches.}
\begin{adjustbox}{width=.5\textwidth,center}
\begin{tabular}{|lllllll|}
\hline
Train &Approach & Test experiment no.&&&& \\
dataset && &&&& \\ 
&&&T1 &T2&T3&T4 \\ \hline
- & Baseline & MAE {[}deg{]} & 11.7 & 12.4 & 14.7 & 12.8 \\
S1 & GHNet & \begin{tabular}[c]{@{}l@{}}MAE {[}deg{]}\\ Improvement \%\end{tabular} & \begin{tabular}[c]{@{}l@{}}8.3\\ 29.1\end{tabular} & \begin{tabular}[c]{@{}l@{}}9.8\\ 20.9\end{tabular} & \begin{tabular}[c]{@{}l@{}}10.1\\ 31.3\end{tabular} & \begin{tabular}[c]{@{}l@{}}9.4\\ 26.5\end{tabular} \\
E3 & GHNet & \begin{tabular}[c]{@{}l@{}}MAE {[}deg{]}\\ Improvement \%\end{tabular} & \begin{tabular}[c]{@{}l@{}}4\\ 65.8\end{tabular} & \begin{tabular}[c]{@{}l@{}}4.8\\ 61.3\end{tabular} & \begin{tabular}[c]{@{}l@{}}4.7\\ 68\end{tabular} & \begin{tabular}[c]{@{}l@{}}4.5\\ 64.8\end{tabular} \\ \hline
\end{tabular}
\label{table:sumarry_results_general}
\end{adjustbox}
\end{table}
\section{Conclusion}\label{sec:pr}
When a vehicle is travelling at low speeds (less than 2m/s) the model-based GNSS heading accuracy is insufficient. To cope with this problem, we proposed GHNet, a deep-learning framework allowing the regressing of the heading angle based on GNSS velocity measurements, at each epoch. We examined two possibilities for training the network: one using a simplified simulation and the second using a NCLT dataset, while the test dataset was always from the NCLT dataset (not present in the train).
Using the simulation dataset, GHNet improves the MAE performance by $27\%$ compared to the baseline approach. When using the NCLT train dataset the improvement was $65\%$. Thus, there exists a trade-off between simplicity—as a simulation dataset does not require extensive field experiments—and accuracy, since using the NCLT train dataset the performance improved by a factor of 2.4. When the vehicle speed is greater than $2 m/s$, the baseline approach outperforms GHNet in its current architecture and obtains satisfactory results. As our goal was to improve low vehicle speeds, no effort was made to improve our architecture to cope with speeds greater than $2 m/s$.\\
GHNet is a shallow network based on 1D-CNN and fully connected layers. As a deep-learning approach it is characterized by its capabilities for noise reduction  and capturing nonlinear behaviour in the system, and yet due to its small size it can be easily implemented for real-time applications. Also, GHNet does not require an additional antenna or hardware; just software modifications.  Our GHNet can be applied to any type of ground vehicles such as passenger cars, autonomous vehicles, autonomous ground vehicles, off-road vehicles, and mobile robots.

\bibliographystyle{IEEEtran}
\bibliography{IEEEabrv,biblio_traps_dynamics,reference}

\begin{thebibliography}{10}
\providecommand{\url}[1]{#1}
\csname url@samestyle\endcsname
\providecommand{\newblock}{\relax}
\providecommand{\bibinfo}[2]{#2}
\providecommand{\BIBentrySTDinterwordspacing}{\spaceskip=0pt\relax}
\providecommand{\BIBentryALTinterwordstretchfactor}{4}
\providecommand{\BIBentryALTinterwordspacing}{\spaceskip=\fontdimen2\font plus
\BIBentryALTinterwordstretchfactor\fontdimen3\font minus
  \fontdimen4\font\relax}
\providecommand{\BIBforeignlanguage}[2]{{%
\expandafter\ifx\csname l@#1\endcsname\relax
\typeout{** WARNING: IEEEtran.bst: No hyphenation pattern has been}%
\typeout{** loaded for the language `#1'. Using the pattern for}%
\typeout{** the default language instead.}%
\else
\language=\csname l@#1\endcsname
\fi
#2}}
\providecommand{\BIBdecl}{\relax}
\BIBdecl

\bibitem{britting2010inertial}
Britting,~K.~R., \emph{{Inertial} navigation systems analysis}, 2010.

\bibitem{titterton2004strapdown}
Titterton,~D., Weston,~J.~L., and Weston,~J., \emph{{Strapdown} inertial
  navigation technology}.\hskip 1em plus 0.5em minus 0.4em\relax IET, 2004,
  vol.~17.

\bibitem{noureldin2012fundamentals}
Noureldin,~A., Karamat,~T.~B., and Georgy,~J., \emph{{Fundamentals} of inertial
  navigation, satellite-based positioning and their integration}.\hskip 1em
  plus 0.5em minus 0.4em\relax Springer Science \& Business Media, 2012.

\bibitem{farrell2008aided}
Farrell,~J., \emph{{Aided} navigation: {GPS} with high rate sensors}.\hskip 1em
  plus 0.5em minus 0.4em\relax McGraw-Hill, Inc., 2008.

\bibitem{klein2010pseudo}
Klein,~I., Filin,~S., and Toledo,~T., ``{Pseudo}-{Measurements} as {Aiding} to
  {INS} during {GPS} {Outages},'' \emph{Navigation}, vol.~57, no.~1, pp.
  25--34, 2010.

\bibitem{engelsman2023information}
Engelsman,~D. and Klein,~I., ``{Information} {Aided} {Navigation}: {A}
  {Review},'' \emph{arXiv preprint arXiv:2301.01114}, 2023.

\bibitem{grewal2007global}
Grewal,~M.~S., Weill,~L.~R., and Andrews,~A.~P., \emph{{Global} positioning
  systems, inertial navigation, and integration}.\hskip 1em plus 0.5em minus
  0.4em\relax John Wiley \& Sons, 2007.

\bibitem{groves2015principles}
Groves,~P.~D., ``{Principles} of {{GNSS}}, inertial, and multisensor integrated
  navigation systems, {[Book} review],'' \emph{IEEE Aerospace and Electronic
  Systems Magazine}, vol.~30, no.~2, pp. 26--27, 2015.

\bibitem{7004783}
Chang,~L., Li,~K., and Hu,~B., ``{Huber’s} {M}-{Estimation}-{Based} {Process}
  {Uncertainty} {Robust} {Filter} for {Integrated} {INS/GPS},'' \emph{IEEE
  Sensors Journal}, vol.~15, no.~6, pp. 3367--3374, 2015.

\bibitem{rhee2004observability}
Rhee,~I., Abdel-Hafez,~M.~F., and Speyer,~J.~L., ``{Observability} of an
  integrated {{GPS/INS}} during maneuvers,'' \emph{IEEE Transactions on
  Aerospace and Electronic systems}, vol.~40, no.~2, pp. 526--535, 2004.

\bibitem{tang2008ins}
Tang,~Y., Wu,~Y., Wu,~M., Wu,~W., Hu,~X., and Shen,~L., ``{INS/GPS}
  integration: {Global} observability analysis,'' \emph{IEEE Transactions on
  Vehicular Technology}, vol.~58, no.~3, pp. 1129--1142, 2008.

\bibitem{seo2006lever}
Seo,~J., Lee,~H.~K., Lee,~J.~G., and Park,~C.~G., ``{Lever} arm compensation
  for {{GPS/INS}}/odometer integrated system,'' \emph{International Journal of
  Control, Automation, and Systems}, vol.~4, no.~2, pp. 247--254, 2006.

\bibitem{borko2018gnss}
Borko,~A., Klein,~I., and Even-Tzur,~G., ``{{GNSS/INS}} fusion with virtual
  lever-arm measurements,'' \emph{Sensors}, vol.~18, no.~7, p. 2228, 2018.

\bibitem{9722837}
Liu,~Y., Luo,~Q., and Zhou,~Y., ``Deep learning-enabled fusion to bridge gps
  outages for ins/gps integrated navigation,'' \emph{IEEE Sensors Journal},
  vol.~22, no.~9, pp. 8974--8985, 2022.

\bibitem{klein2011modified}
Klein,~I., Filin,~S., and Toledo,~T., ``{A} modified loosely coupled approach
  to {INS/GPS} integration,'' 2011.

\bibitem{Pereira2016GPSHA}
\BIBentryALTinterwordspacing
Pereira,~R., ``{GPS} {Heading} and {Pitch} {Estimation} using
  {Single}-{Frequency} , {Dual}-{Frequency} or {Wide}-{Lane} {Measurements},''
  2016. [Online]. Available:
  \url{https://api.semanticscholar.org/CorpusID:202549107}
\BIBentrySTDinterwordspacing

\bibitem{wang2019research}
Wang,~H., Liu,~N., Su,~Z., and Li,~Q., ``{Research} on low-cost attitude
  estimation for {MINS}/dual-antenna {GNSS} integrated navigation method,''
  \emph{Micromachines}, vol.~10, no.~6, p. 362, 2019.

\bibitem{roth2012improving}
Roth,~J., Kaschwich,~C., and Trommer,~G., ``{Improving} {GNSS} attitude
  determination using inertial and magnetic field sensors,'' \emph{Inside
  GNSS}, vol.~7, no.~1, pp. 54--62, 2012.

\bibitem{petovello2017you}
Petovello,~M., ``{How} do you use {GNSS} to compute the attitude of an
  object,'' \emph{Insid GNSS}, 2017.

\bibitem{pereira2016performance}
Pereira,~R. and Sanguino,~J., ``{Performance} evaluation of the ambiguity
  filter as an alternative of using dual-frequency measurements for {GPS}
  heading and pitch estimation,'' in \emph{2016 European Navigation Conference
  (ENC)}.\hskip 1em plus 0.5em minus 0.4em\relax IEEE, 2016, pp. 1--10.

\bibitem{shin2002accuracy}
Shin,~E.-H. and El-Sheimy,~N., ``{Accuracy} improvement of low cost {INS/GPS}
  for land applications,'' in \emph{Proceedings of the 2002 national technical
  meeting of the institute of navigation}, 2002, pp. 146--157.

\bibitem{salycheva2004kinematic}
Salycheva,~A. and Cannon,~M., ``{Kinematic} azimuth alignment of {INS} using
  {GPS} velocity information,'' in \emph{Proceedings of the 2004 National
  Technical Meeting of The Institute of Navigation}, 2004, pp. 1103--1113.

\bibitem{godha2006performance}
Godha,~S., ``{Performance} evaluation of low cost {MEMS}-based {IMU} integrated
  with {GPS} for land vehicle navigation application,'' \emph{UCGE report}, no.
  20239, 2006.

\bibitem{9069956}
Klein,~I., Lipman,~Y., and Vaknin,~E., ``{Squeezing} {Position} {Updates} for
  {Enhanced} {Estimation} of {Land} {Vehicles} {Aided} {INS},'' \emph{IEEE
  Sensors Journal}, vol.~20, no.~16, pp. 9385--9393, 2020.

\bibitem{klein2022data}
Klein,~I., ``{Data}-driven meets navigation: {Concepts}, models, and
  experimental validation,'' in \emph{2022 DGON Inertial Sensors and Systems
  (ISS)}.\hskip 1em plus 0.5em minus 0.4em\relax IEEE, 2022, pp. 1--21.

\bibitem{cohen2023inertial}
Cohen,~N. and Klein,~I., ``{Inertial} {Navigation} {Meets} {Deep} {Learning}:
  {A} {Survey} of {Current} {Trends} and {Future} {Directions},'' \emph{arXiv
  preprint arXiv:2307.00014}, 2023.

\bibitem{base_navigation_equation}
Salycheva,~A. and Cannon,~M., ``{Kinematic} azimuth alignment of {INS} using
  {GPS} velocity information,'' in \emph{Proceedings of the 2004 National
  Technical Meeting of The Institute of Navigation}, 2004, pp. 1103--1113.

\bibitem{Batch1}
Bjorck,~N., Gomes,~C.~P., Selman,~B., and Weinberger,~K.~Q., ``{Understanding}
  batch normalization,'' \emph{Advances in neural information processing
  systems}, vol.~31, 2018.

\bibitem{Batch2}
Zhang,~P., Zhang,~L., Leung,~H., and Wang,~J., ``{A} deep-learning based
  precipitation forecasting approach using multiple environmental factors,'' in
  \emph{2017 IEEE International Congress on Big Data (BigData Congress)}.\hskip
  1em plus 0.5em minus 0.4em\relax IEEE, 2017, pp. 193--200.

\bibitem{Gelus2}
Hendrycks,~D. and Gimpel,~K., ``{Gaussian} error linear units (gelus),''
  \emph{arXiv preprint arXiv:1606.08415}, 2016.

\bibitem{Gelus3}
Yu,~C. and Su,~Z., ``{Symmetrical} gaussian error linear units {(sgelus)},''
  \emph{arXiv preprint arXiv:1911.03925}, 2019.

\bibitem{Droput2}
Srivastava,~N., Hinton,~G., Krizhevsky,~A., Sutskever,~I., and
  Salakhutdinov,~R., ``{Dropout}: a simple way to prevent neural networks from
  overfitting,'' \emph{The journal of machine learning research}, vol.~15,
  no.~1, pp. 1929--1958, 2014.

\bibitem{adamW}
Loshchilov,~I. and Hutter,~F., ``{Decoupled} weight decay regularization,''
  \emph{arXiv preprint arXiv:1711.05101}, 2017.

\bibitem{NCLT}
Carlevaris-Bianco,~N., Ushani,~A.~K., and Eustice,~R.~M., ``{University} of
  {Michigan} {North} {Campus} long-term vision and lidar dataset,'' \emph{The
  International Journal of Robotics Research}, vol.~35, no.~9, pp. 1023--1035,
  2016.

\end{thebibliography}

\end{document}